\documentclass{article}

\usepackage{microtype}
\usepackage{graphicx}
\usepackage{subfigure}
\usepackage{booktabs} 
\usepackage{url}

\usepackage{hyperref}

\usepackage[accepted]{icml2019}

\icmltitlerunning{Personalized Student Stress Prediction with Deep Multitask Network}

\begin{document}

\twocolumn[
\icmltitle{Personalized Student Stress Prediction with Deep Multitask Network}



\icmlsetsymbol{equal}{*}

\begin{icmlauthorlist}
\icmlauthor{Abhinav Shaw}{equal,to}
\icmlauthor{Natcha Simsiri}{equal,to}
\icmlauthor{Iman Deznabi}{to}
\icmlauthor{Madalina Fiterau}{to}
\icmlauthor{Tauhidur Rahman}{to}
\end{icmlauthorlist}

\icmlcorrespondingauthor{Abhinav Shaw}{abhinavshaw@umass.edu, abhinav.shaw1993@gmail.com}
\icmlcorrespondingauthor{Natcha Simsiri}{nsimsiri@umass.edu}
\icmlaffiliation{to}{College of Computer Science, University of Massachusetts, Amherst, USA}

\icmlkeywords{Stress Prediction, Neural Networks, Multi-Task Learning}

\vskip 0.3in
]



\printAffiliationsAndNotice{\icmlEqualContribution} 

\begin{abstract}

With the growing popularity of wearable devices, the ability to utilize physiological data collected from these devices to predict the wearer's mental state such as mood and stress suggests great clinical applications, yet such a task is extremely challenging. In this paper, we present a general platform for personalized predictive modeling of behavioural states like students' level of stress. Through the use of Auto-encoders and Multitask learning we extend the prediction of stress to both sequences of passive sensor data and high-level covariates. Our model outperforms the state-of-the-art in the prediction of stress level from mobile sensor data, obtaining a 45.6\% improvement in F1 score on the StudentLife dataset.

\end{abstract}
\vspace{-0.2in}
\section{Introduction}
\label{introduction}

Today's competitive and demanding environment often overwhelms students with assignments, tests and part-time work. A prolonged exposure to stressful academic and social environment causes cardiovascular diseases \cite{Rozanski, Kario}, alterations of the brain causing differences in memory and cognition \cite{Lupien}, suppression of the immune system \cite{Khansari}, and poor academic performance \cite{Akane2, MickyT}.
 With the efforts of researchers at various institutions several technologies for detecting stress has been accomplished. Few use heart rate and heart rate variability \cite{Vrijkotte}, cortisol levels \cite{Dickerson} and skin conductance \cite{Setz}. Other techniques do not depend on sensors but simply try to discover the user's stress through self-reporting tools e.g., \cite{rahman2014towards} and surveys like the Perceived Stress Scale \cite{Cohen}. 
 
  With the induction of high quality, robust sensors in wearables like FitBit, Apple Watch and smartphones, efficient collection of physiological and behavioural data with reasonable accuracy has become affordable. The StudentLife study in \cite{Studentlife} collected Sleep Patterns, Activity, Conversation, Location, information regarding Mental Health like stress levels and much more through the StudentLife application on android smartphones.
  
  Contemporary research such as \cite{Akane1} and \cite{Akane2} has leveraged similar type of data from sensors and Machine Learning to predict stress levels of students. Furthermore, using the data collected from the StudentLife application, \cite{Rwang} have been successful in classifying students' as depressed or not in a binary classification problem. 
  However, the task of predicting human psychological state (e.g., stress) using passive sensing data on a multi-class classification problem remains a challenge. Lack of gold standard labels, noisy raw sensor data, heterogeneity in granularity and inter subject variability in behavioural and environmental patterns have stymied predictive modeling of this kind.
  
  \cite{Gatis} have tried to predict stress of students in the StudentLife dataset by novel feature engineering of location based features and Neural Networks. To the best of our knowledge their model is the state-of-the-art on predictive modeling of stress on this dataset and we refer this model as Location Based MultiLayer Perceptron (Location-MLP). However, it doesn't address the challenges of inter-subject variability or heterogeneity in granularity. The model is also limited to location and few covariates as features.
  In this paper we introduce the \textbf{Cross-personal Activity LSTM Multitask Auto-encoder Network (CALM-Net)} which considers data as time-series and is able to identify temporal patterns contained in student data. By including these different levels of information and personalizing the predictions to students, CALM-Net can achieve an average \textit{F1-score} of \textbf{0.594}  which is an improvement of \textbf{45.6\%} when compared to the Location Based MLP.
  CALM-Net offers the flexibility to personalize models and the ability to incorporate time-series information, which in general, can be used by researchers to improve performance for categorical prediction of psychological states of humans.
   
\section{Background Information}

 The StudentLife study was conducted in Dartmouth college where passive sensing and survey data was collected over 10 weeks among 48 students. The data collection, which mainly comprises of sleep patterns, activity, meal counts and conversations was facilitated by the StudentLife application on a smartphone. On a daily basis the StudentLife application collected stress data on a scale of 1-5 in the form of Ecological Momentary Assessments (EMA) which are responses to questionnaires in real time. Although EMAs are self-reported and consequently noisy, they are usually a good indicator of the actual stress state of the person making it feasible but not ideal to use them as labels for supervised learning tasks. 
 
 Out of the several challenges in the predictive modeling of stress, one major challenge that makes this task formidable is that the features have disparate granularity; they are also missing at random, which could be caused by technical issues such as a sensor failure or the phone being switched off. The dataset is heterogeneous in nature since the data collected, is from a variety of sources like passive sensors, surveys and self reported EMAs. Out of the many discrete sequence and covariate features in StudentLife dataset, we select the ones that suggest evidence of these being good predictors of stress in \cite{Matthew, Akane1, MickyT, Akane2} etc.
 
Among the discrete sequence data, we use Activity and Audio which are categorical integer values, recorded by the StudentLife app as follows \textit{0- No Activity/Silence, 1- Walking/Voice, 2- Running/Noise, 3- Unknown}. Conversation, Phone Charge, Phone Lock are all inferred as binary values. These features are recorded at a variable rate, ranging from once every 10 seconds to once every minute.

Along with the above passive sensing data our model uses inferred and recorded covariates: Day of Week, Sleep Rating, Sleep duration (all recorded or inferred as integer values) and a binary covariate ``Exam Period''. We use time to next deadline as a feature which is inferred by the recorded deadlines in StudentLife. We believe that as a deadline approaches the stress levels of students must increase. The features, covariates, their respective value bounds and modes are listed in appendix Table \ref{tab:feat_details}.

\section{Methods}

\subsection{Problem Setup}
Passive sensors in smartphones allow collection of rich discrete time-series data crucial for stress prediction. The raw and elongated time-series features such as Activity and Audio can contribute to a few thousand data points everyday and cannot be used for training `as is' due to infrequent label samples.
To deal with this we first bin the whole time-series into 1-minute bins and take the mode of the categorical inferences. This offers a compact representation of what the subject was doing in that minute, for example \textit{`was the wearer running or having a conversation?'}. This results in 1,440 sequences per day, which is still a very long sequence to be modeled using Recurrent Neural Networks. We further address this by computing the histogram of the features in 1-hour bins yielding 24 sequences per stress label. Intuitively, we are modeling how much conversation or activity a student has undergone in an hour which led to the stress label in a consistent manner, removing any bias due to irregular sampling of sensors by containing that information in a base bin of 1-minute. This type of feature engineering can easily be extended to different datasets with similar/same type of passive sensing time-series data as the final histogram is independent of the initial binning granularity of 1 min and can accommodate even higher-resolution data.


\subsection{Models}
\label{section-3.2}
\begin{figure*}[ht]
\begin{center}
\centerline{\includegraphics[width=0.80\textwidth]{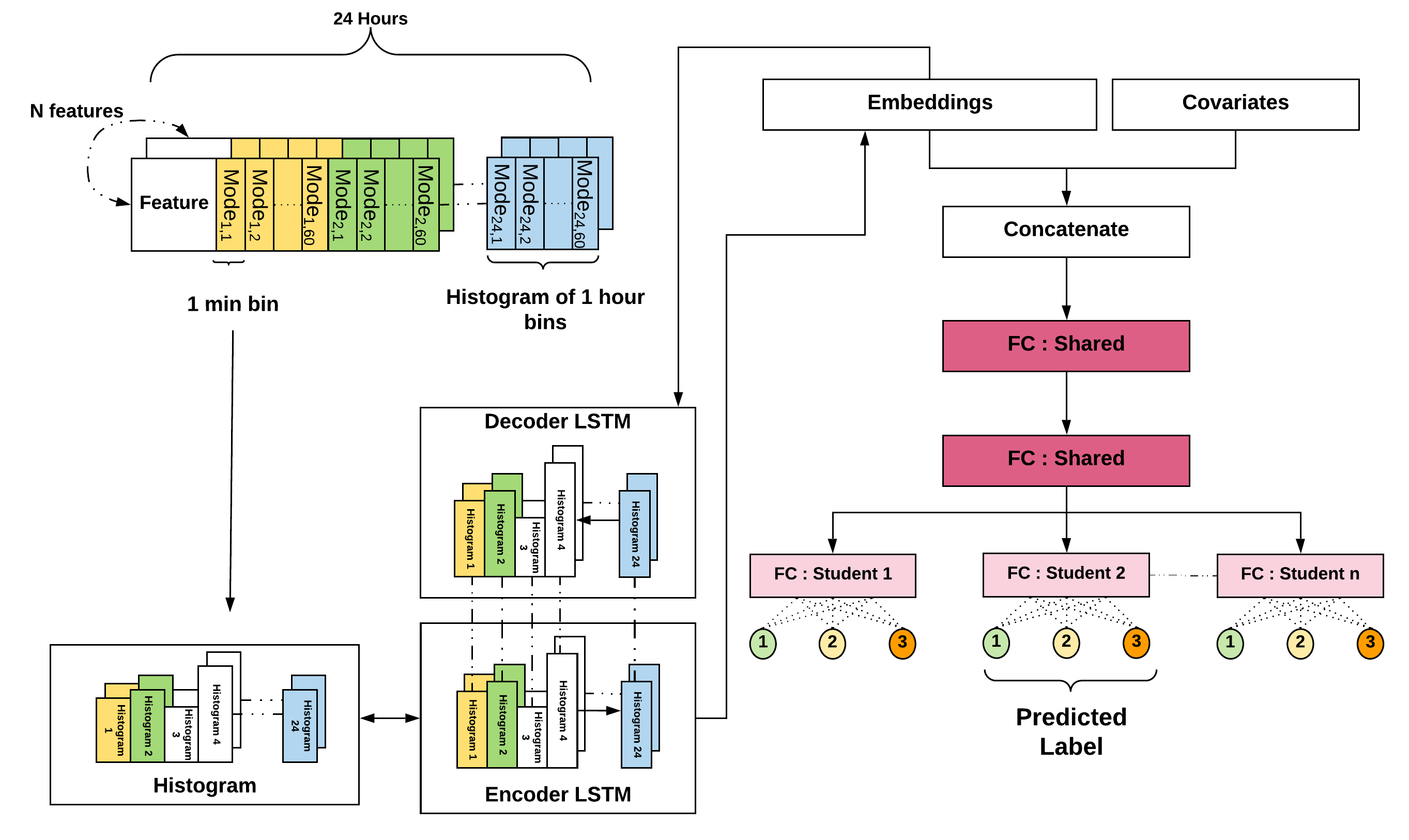}}
\caption{Cross-personal Activity LSTM Multitask Auto-encoder Network (CALM-Net).}
\label{model}
\end{center}
\vskip -0.3in
\end{figure*}

\subsubsection{Location Feature Based MLP}

In the work done by \cite{Gatis}, a Multilayer Perceptron (MLP) with 4 fully connected layers was employed to perform stress inference. Each fully connected layer uses the \textit{tanh} activation followed by a Batch Normalization and Dropout layer. The input to the model is feature engineered GPS data aggregated on a daily basis. There are a total of 8 location features and 4 covariates. 
The location features are total distance covered, max displacement, distance entropy on 10 minutes bins, distance standard deviation, number of unique tiles visited, difference in tiles visited from the previous day, approximate area of the GPS convex hull, and number of clusters on the GPS data. Tiles are non-overlapping, consecutively partitioned squares of 50 meters on the sides, where the combined squares represent the area the student may have traversed to. Each tile is uniquely labeled and counts on tiles is one of the features used. Throughout a day, the sequence of tiles were visited, and the difference in sequence of tiles covered compared to one on the previous were computed using Levenshtein edit-distance. Finally, the covariates used are indicators of whether the day is the start of term, mid-term, end of term or a weekend. 

We followed the paper in building the baseline and also achieve an F-1 score of \textit{0.42} with their experiment setup. To the best of our knowledge this is the-state-of-the-art on the StudentLife dataset. The features engineered with this baseline was done to the best of our abilities, although we note there may have been some discrepancy with the original work in obtaining certain features.



\subsubsection{LSTM}


The state-of-the-art model which utilizes featured engineered aggregates doesn't model the time-series. This leads to an inability to use the information in granular passive sensing data which is ubiquitous in these kinds of datasets. To model the temporal patterns of features like Activity, Audio and Conversation we put the sequences of hourly histograms through an LSTM (Long Short-term Memory Network) \cite{lstm}. Then we concatenate the last hidden state with the covariates. This concatenated output is passed through multiple layers of fully connected layers with the ReLU activation to finally obtain the class probabilities by using Weighted Categorical Cross-Entropy. This as one of our baselines and compare our final method to this. It is also a part of our next model.

\subsubsection{LSTM Multitask Network (LM-Net)}

Due to the heterogeneous nature of the dataset it is hard to incorporate information at different levels of granularity for a predictive task. Furthermore, trying to capture personal dynamics of all subjects using one model is demanding, as these dynamics are very distinct and have high inter-subject variability.
To learn personalized models for each student, we follow \cite{DBLP:conf/ijcai/JaquesRTSP17} and use a Multitask approach which comprises of a LSTM to model sequence of histograms followed by shared fully connected layers and a MLP for each student. A similar approach was also taken in \cite{Kandemir} for the prediction of affect (mood) by learning user specific kernels.
As indicated by our experiments detailed in section \ref{results_section}, this approach can learn the differences between students and subsequently yield significant improvement in performance. It also gives evidence that learning a single model for all the students is unsuitable. Multitask learning also acts as a heavy regularizer, preventing the model to overfit for one student or the most common label. The shared layers learn common features, while the personal layers learn features that are relevant to the respective subject.





\subsubsection{Cross-personal Activity LSTM Multitask Auto-encoder Network (CALM-Net)}

Amongst the popular techniques for modeling time-series data, variations of RNNs like GRU and LSTM are the most popular, however people have used Auto-encoders for compression and reduction of the temporal dimensions. In \cite{LANGKVIST201411}, different techniques for time-series have been summarised. Out of which we try RNN-LSTM and Auto-encoders.
In CALM-Net we replace the LSTM layer with an LSTM Auto-encoder. Due to the low amount of training data available, we find it useful to reconstruct the sequence of histograms through an Encoder-Decoder pair. This also ensures that the model does not overfit to the discrete training sequences.
For calculating reconstruction error between the decoded sequences and the original sequences we used Mean Absolute Error(MAE). The final error/loss (expression in equation \ref{equation-3}) is a weighted sum of the Reconstruction Error and Classification Error where $\alpha$ and $\beta$ are hyperparameters.

  \begin{equation} 
    \label{equation-3}
    \alpha * RE + \beta * CE
  \end{equation}


\section{Result}

\label{results_section}
Due to a heavy imbalance of class labels on a scale of 1-5, we follow \cite{Gatis}, converting the five stress label scale to a scale of three stress labels by defining our classes as - \textit{below median stress, median stress and above median stress}. 
We determined that some students present have high level of inter-subject variability which will make classification of stress extremely difficult for the current methods and selected the students who have greater than 40 labels and trained CALM-Net with a learning rate of $10^{-6}$ and a weight decay of $10^{-4}$. The details of the data split and model configuration is given in appendix.


To evaluate our methods and make a fair comparison with baselines and previous state-of-the-art. We report the average F1-score achieved by each method on 5-fold cross validation. 
We compare against the state-of-the-art Location-MLP method with the data of a full day on which the label was reported, which potentially uses some data that is recorded after the stress label. Furthermore, we are comparing against LSTM, LSTM Multitask Network (LM-Net) which are explained in section \ref{section-3.2}. The achieved results are summarised in Table~\ref{results-table}.
As you can see, considering the data as time-series along with additional features available in the dataset improves the performance of the model as indicated by LSTM model improving upon the performance of the previous state-of-the-art model (Location-MLP). Since we have personalized models for every student, we also considered another baseline which is just predicting the most common label for the respective student as their stress state and outperforming it. From the results it is evident that personalizing the model to students can outperform state-of-the-art model (Location-MLP) and LSTM, showing the value of personalizing the methods to students. You can further see that CALM-Net with a \textit{F1-score} of 0.594 outperforms the other models due to its ability to capture both temporal patterns and learning personalized information about the students. The detailed model configuration is given in appendix section \ref{sec:conf}. 
\cite{amtl_8_2019}


\begin{table}[t]
\caption{F1 scores of stress level prediction on StudentLife dataset.}
\label{results-table}
\begin{center}
\begin{small}
\begin{tabular}{lcccr}
\toprule
Model & F1-score\\
\midrule
Location-MLP & 0.408 \\ 
LSTM & 0.426\\ 
Most Common Label & 0.551\\
LM-Net & 0.586 \\
CALM-Net No covariates  & 0.571\\
\textbf{CALM-Net } & \textbf{0.594} \\
\bottomrule
\end{tabular}
\end{small}
\end{center}
\vskip -0.2in
\end{table}

\begin{table}[t]
\centering
\caption{Percent gain in F1 Score with Multitask Learning compared to the LSTM model on varying number of students.}
\label{comparision-table} 
\begin{center}
\begin{small}
\begin{tabular}{lcccr}
\toprule
Students & F1-score & F1-score & \% Inc \\
 & w/o Multitask & Multitask &  \\
\midrule
5 & 0.47 & 0.585 & 24.2 \% \\
13 & 0.436 & 0.583 & 33.7 \% \\
23 & 0.426 & 0.594 & 39.4 \% \\
\bottomrule
\end{tabular}
\end{small}
\end{center}
\vspace{-0.2in}
\end{table}

Since CALM-Net can learn personalized patterns for each student it yields better performance as we increase the number of students.
To test this hypothesis we designed an experiment where we try our model without Multitask heads and with Multitask heads on 5 ,13 and 23 students. The results of this experiment are summarised in Table~\ref{comparision-table}. These results indicate that when we increase the number of students the performance of the model without Multitask heads will drops significantly, while the model with Multitask heads will achieve almost the same or better results, validating our hypothesis. 


\section{Discussion}

CALM-Net yields superior performance as it is able to model the temporal events contributing to stress of a subject while dealing with long sequences of sensor data. The Auto-encoder prevents the model to overfit on the training sequences and provides an additional boost to the \textit{F1-score}.
The ability of CALM-Net to incorporate granular temporal information and high-level covariates, along with an architecture which is capable of deciphering personalized patterns for each student without overfitting, contributes to its high performance. Multitask learning improves the performance of all evaluated models, showing that stress indicators can generally be better modeled using personalized layers.

\section{Conclusion}
We presented CALM-Net model for predicting stress levels in StudentLife dataset.
Our models are specially designed to solve three challenges which are ubiquitous in passive sensing datasets. First, it presents a general platform to address the issue of data heterogeneity with use of LSTM Auto-encoders.
Second, it is able to deal with long and irregular sequences by feature engineering and histogram of categorical inference values addressing the Multi-Resolution nature of the data which is commonplace in the field. 
Third, by creating personalized models for every student while leveraging information from all the students it is able to achieve a \textit{F1-score} of \textbf{0.594}. This allows us to cope with inter-subject variability providing significant improvement upon previous state-of-the-art models. 
We note that while the model performs well on given set of students, it needs some data for every student to be able to train their respective MLPs, so the model is unable to predict stress level of new students, we leave addressing this limitation for future work.

\section{Acknowledgement}

We would like to thank Nick Monath, Rasmus Lundsgaard Christiansen and Professor Andrew McCallum for the initial opportunity to work on the StudentLife dataset. Abhinav Shaw and Natcha Simsiri were supported by the Center of Data Science at University of Massachusetts Amherst.

\nocite{langley00}

\bibliography{references.bib}
\bibliographystyle{icml2019}
\clearpage

\appendix

\onecolumn
\section{Appendix}
\label{Appendix}

  
\subsection{Used Features}
\label{features-used}
In all, our data comprises of 23 students, totaling to 1183 data points achieving roughly equal amount of training data in \cite{Gatis}. These 1183 data points have the following label distribution - 263 below median stress, 511 median stress and 409 above median. Since student 59 has 269 labels which is on an average, four times the number of labels of other students, is removed from the training set as he/she may dominate the shared layer and skew our predictions. 

The list of student IDs used for training - [4, 7, 8, 10, 14, 16, 17, 19, 22, 23, 24, 32, 33, 35, 36, 43, 44, 49, 51, 52, 53, 57, 58] 

We use both time-serie features and covariates as input to our LSTM, LM-Net and CALM-Net models. The time-serie features we used are time to next label, time to next deadline, activity mode (a discrete scale of 0 to 3 indicating being sedentary to high level of activity such as running), conversation duration mode, phone charge duration mode, and phone lock duration mode. Finally the covariates used are the day of the week, sleep rating, hours slept and an indicator for an exam period. 

\begin{table}[h!]
    \centering
    \begin{tabular}{lcccr}
        \toprule
        \textbf{Feature Type} & \textbf{Feature Name} & \textbf{Feature Values}  & \textbf{Mode in dataset} \\
        \midrule
        \textbf{Discrete Sequence} & Activity & $[0,3]$ & 0 \\
        & Audio & $[0,3]$  & 0 \\
        & Conversation & $[0,1]$ & 0 \\
        & Phone Charge & $[0,1]$ & 0 \\
        & Phone Lock & $[0,1]$ & 1 \\
        & Day of the Week & $[0, 6]$ & N/A \\
        & Exam Period & $[0, 1]$ & 0 \\
        \textbf{Covariates} & Time to next deadline & $[0, \infty+)$ & N/A \\
        & Sleep Rating & $[0, \infty+)$ & N/A \\
        & Sleep Duration & $[0, \infty+)$ & N/A \\
        
        \bottomrule
    \end{tabular}
    \caption{Here we list the features we used in our experiment}
    \label{tab:feat_details}
\end{table}

\subsection{Model Configurations}
\label{sec:conf}
\begin{table}[h!]
    \centering
    \begin{tabular}{lr}
        \toprule
        \textbf{Hyper-parameter} & \textbf{value} \\
        \midrule
        $\alpha$ & 0.001\\
        $\beta$ & 1\\
        Auto-encoder embedding size & 128 \\
        shared layers hidden size & 256 \\
        Personal layer hidden size & 64\\
        \bottomrule
    \end{tabular}
    \caption{The configuration details of CALM-NET}
    \label{tab:conf_details}
\end{table}


\end{document}